\documentclass[10pt,twocolumn,letterpaper]{article}

\usepackage{cvpr}              
\usepackage{tabularx}
\usepackage{multirow}
\newcolumntype{Y}{>{\centering\arraybackslash}X}
\usepackage{graphicx}
\usepackage{amssymb,pifont}
\newcommand{\cmark}{\ding{51}}  
\definecolor{cvprblue}{rgb}{0.21,0.49,0.74}
\usepackage[pagebackref,breaklinks,colorlinks,allcolors=cvprblue]{hyperref}

\def\paperID{***} 
\def\confName{CVPR}
\def\confYear{2026}

\title{HeroGS: Hierarchical Guidance for Robust 3D Gaussian Splatting under Sparse Views}

\author{Jiashu Li$^{\dagger}$\quad Xumeng Han$^{\dagger}$\quad Zhaoyang Wei$^{\dagger}$\quad
Zipeng Wang\quad Kuiran Wang \quad Guorong Li \\
\quad Zhenjun Han$^{*}$\quad Jianbin Jiao\\
University of the Chinese Academy of Sciences\\
{\tt\small lijiashu24@mails.ucas.ac.cn}
}

\begin{document}
\maketitle
\let\thefootnote\relax\footnotetext{$\dagger$ Equal contribution.}
\let\thefootnote\relax\footnotetext{$*$ Corresponding author.}
\begin{abstract}
3D Gaussian Splatting (3DGS) has recently emerged as a promising approach in novel view synthesis, combining photorealistic rendering with real-time efficiency. However, its success heavily relies on dense camera coverage; under sparse-view conditions, insufficient supervision leads to irregular Gaussian distributions—characterized by globally sparse coverage, blurred background, and distorted high-frequency areas.
To address this, we propose \textbf{HeroGS}—\textit{\textbf{H}i\textbf{e}rarchical Guidance for \textbf{Ro}bust 3D \textbf{G}aussian \textbf{S}platting}—a unified framework that establishes hierarchical guidance across the image, feature, and parameter levels. 
At the image level, sparse supervision is converted into pseudo-dense guidance, globally regularizing the Gaussian distributions and forming a consistent foundation for subsequent optimization. 
Building upon this, Feature-Adaptive Densification and Pruning (FADP) at the feature level leverages low-level features to refine high-frequency details and adaptively densifies Gaussians in background regions.
The optimized distributions then support Co-Pruned Geometry Consistency (CPG) at parameter level, which guides geometric consistency through parameter freezing and  co-pruning, effectively removing inconsistent splats. 
The hierarchical guidance strategy effectively constrains and optimizes the overall Gaussian distributions, thereby enhancing both structural fidelity and rendering quality.
Extensive experiments demonstrate that \textbf{HeroGS} achieves  high-fidelity reconstructions and consistently surpasses state-of-the-art baselines under sparse-view conditions.
\end{abstract}    
\section{Introduction}
\label{sec:intro}

Reconstructing high-fidelity 3D scenes for photorealistic novel view synthesis remains a core challenge in computer vision. While Neural Radiance Fields (NeRF)~\cite{mildenhall2021nerf} and its variants~\cite{barron2021mipnerf,Barron_2022_CVPR,barron2023zipnerf} achieve remarkable visual quality, their implicit representation incurs slow rendering. Recently, 3D Gaussian Splatting (3DGS)~\cite{kerbl20233d} introduces an explicit and efficient alternative, delivering NeRF-level fidelity with real-time performance, and a series of its extensions~\cite{Yu2023MipSplatting,sabourgoli2024spotlesssplats,seo2025flod} rapidly establishing a new state-of-the-art for high-quality, efficient scene reconstruction.

With the success of these methods in dense-view scenarios, the research frontier is now shifting toward the more challenging sparse-input setting. In this regime, the scarcity of viewpoints often leads to irregular Gaussian distributions, resulting in geometric ambiguities and severe rendering artifacts. 
FSGS~\cite{zhu2024fsgs} accelerates early-stage densification to compensate for the insufficient initial Gaussians, while DropGaussian~\cite{Park_2025_CVPR} adopts a dropout-based strategy to regularize the Gaussian distributions in occluded regions. 
However, the absence of comprehensive guidance still leaves the Gaussian field imperfectly optimized, resulting in inaccurate and uneven distributions — with insufficient Gaussians in background regions leading to blurriness, and inadequate supervision over high-frequency details causing misaligned or misplaced Gaussians.
This motivates us to introduce a multi-level framework that globally-to-locally coherent guidance to more comprehensively direct the formation of accurate Gaussian distributions.

\begin{figure}[t]
  \centering
  \includegraphics[width=0.95\columnwidth]{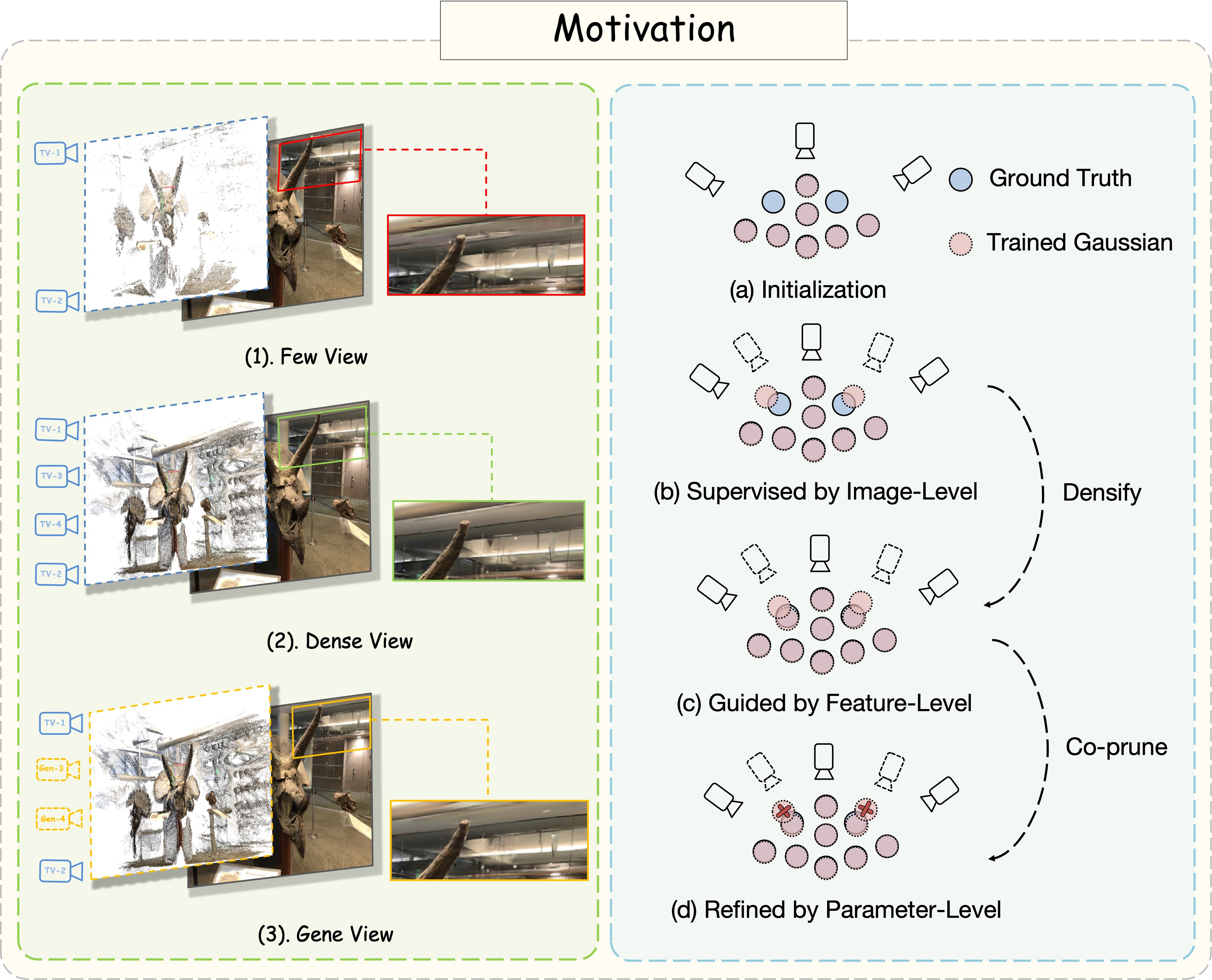}

  \caption{ 
  \textbf{Motivation of HeroGS.} 
  (Left) Image-level pseudo-labels bridge the gap between sparse and dense supervision, yielding a more complete Gaussian distributions. 
  (Right) FADP and CPG refine inaccurate Gaussians by enhancing distributions closer to ground-truth geometry and pruning inconsistencies. 
 }
  \vspace{-10pt}
  \label{fig:new_motivation}
\end{figure}

Sparse-view settings often cause Gaussians to fall outside the field of view, receiving limited gradient feedback and overfitting to a few training images~\cite{Park_2025_CVPR}.
As shown in Fig.~\ref{fig:new_motivation}, adding more views improves gradient coverage and alleviates this issue.
Inspired by this, we synthesize pseudo-labels that, together with real views, provide dense image-level supervision to regularize Gaussian distributions.
This enhanced supervision enriches gradient propagation across the scene, leading to more uniform Gaussian distributions and improved reconstruction fidelity under sparse inputs, while providing richer spatial structural cues for fine-grained guidance at the feature level.

To further refine and regularize the Gaussian distributions in high-frequency regions, we introduce the \textbf{F}eature-\textbf{A}daptive \textbf{D}ensification and \textbf{P}runing (\textbf{FADP}) at the feature level.
FADP increases Gaussian density along edge-aware regions to capture high-frequency details, prunes redundant splats within homogeneous patches to avoid over-saturation, and adds new Gaussians in sparse background areas to improve coverage.
The refined Gaussian distributions from FADP provide a stable foundation for the parameter-level, enabling more reliable co-pruning based on geometric consistency.

At the parameter level, we introduce the \textbf{C}o-\textbf{P}runed \textbf{G}eometry Consistency (\textbf{CPG}), which eliminates abnormal or inconsistent Gaussian distributions via a co-pruning mechanism combined with a post-freeze behavior strategy.
Co-Pruning leverages the self-consistency of the Gaussian field to jointly evaluate and filter Gaussian parameters, retaining only those that exhibit stable and geometrically consistent distributions.
This process effectively preserves robust Gaussians while eliminating unstable or redundant ones, leading to a cleaner and more reliable scene representation, as illustrated in the right part of Fig.~\ref{fig:new_motivation}.
The whole pipeline constitutes a hierarchical guidance strategy that collaboratively optimizes the Gaussian distributions across levels, enhancing its global consistency and fidelity.
Our contributions can be summarized as follows:

\begin{itemize}
   \item  We propose \textbf{HeroGS}, a hierarchical guidance framework that enables compact and high-fidelity 3D reconstructions under sparse-view conditions.
   \item At image level, pseudo-labels are generated to promote more accurate Gaussian distributions. At  feature and parameter levels, the proposed \textbf{FADP} and \textbf{CPG} further refine Gaussian field by enhancing feature-aware density and enforcing geometric consistency, respectively.

   \item Extensive qualitative and quantitative experiments on real-world datasets, including large-scale scenes and various training views, demonstrate that our method significantly outperforms state-of-the-art baselines.
\end{itemize}

\section{Related Work}
\begin{figure*}[t]
  \centering
  \includegraphics[width=0.95\textwidth]{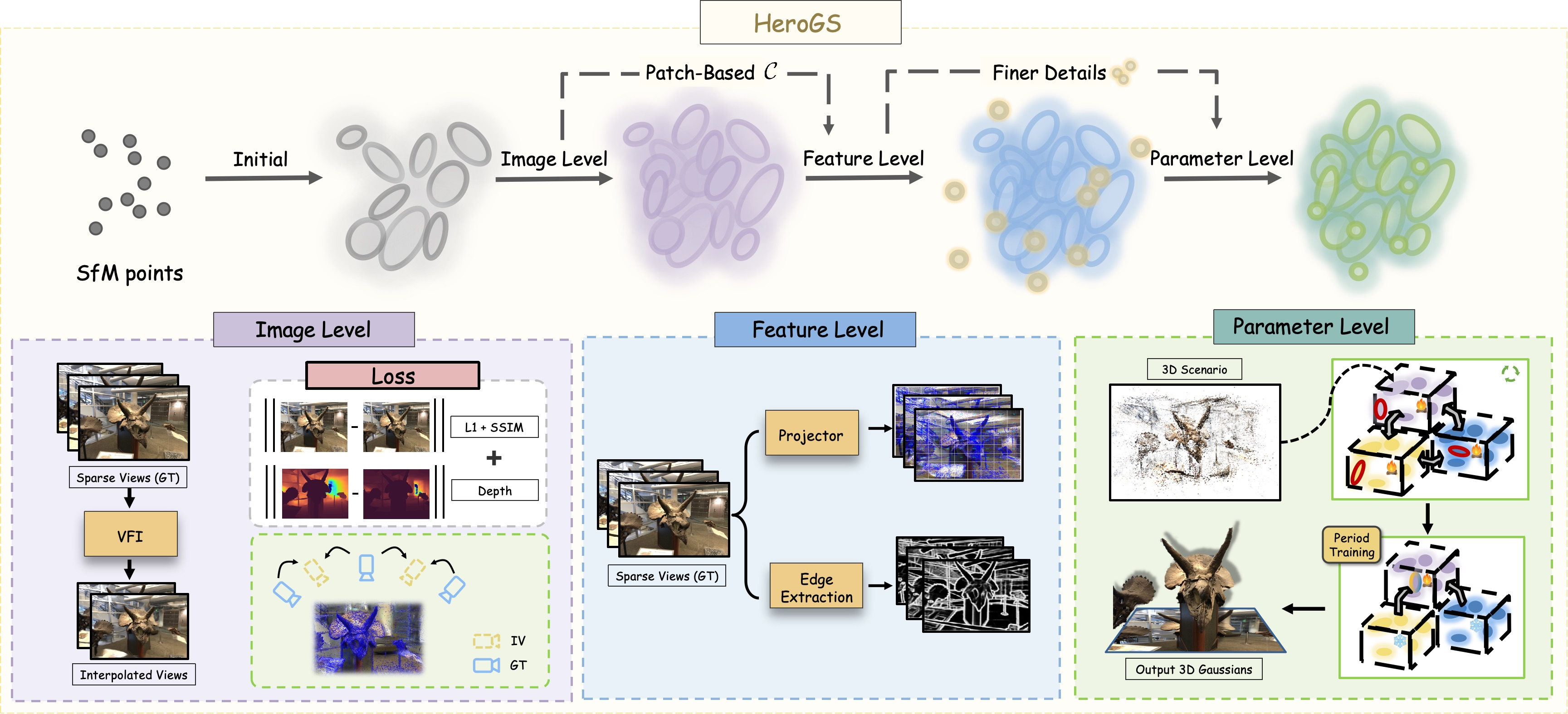}
  \caption{ \textbf{HeroGS Overview.} Initialized from SfM, our framework operates across three levels.
(1) \textbf{Image-level Guidance:} A set of intermediate RGB frames is synthesized from sparse inputs, offering pseudo-dense guidance that globally regularizes the Gaussian distributions and, at the feature level, manifests as patch-based Gaussian numbers $\mathcal{C}$.
(2) \textbf{Feature-level Refinement:} The Feature-Adaptive Densification and Pruning (FADP) leverages edge- and patch-aware features from training views to enhance high-frequency and background regions, while suppressing redundant Gaussians and dilivering finer details for next level.
(3) \textbf{Parameter-level Consistency:} The Co-Pruned Geometry Consistency (CPG) employs auxiliary Gaussian fields with partially frozen parameters to perform co-pruning, eliminating geometrically inconsistent splats.
These levels form a hierarchical guidance with interconnections (dashed lines) that jointly constrains and optimizes the Gaussian field for improved structural fidelity.
 }
  \vspace{-5pt}
  \label{fig:framework}
\end{figure*}
\textbf{Novel View Synthesis.}
\textbf{Neural Radiance Fields (NeRF)}~\cite{mildenhall2021nerf, barron2022mip, bao2023and} and \textbf{3D Gaussian Splatting (3DGS)}~\cite{kerbl20233d} have emerged as two prominent paradigms for high-fidelity novel view synthesis. NeRF-based methods have achieved impressive results in producing photorealistic novel views, especially when abundant views (often hundreds) are available. Meanwhile, 3DGS represents scenes explicitly as a set of 3D Gaussian primitives, which enables faster convergence and real-time rendering performance through GPU-friendly splatting operations. Recent works such as MiniSplatting~\cite{kaiser2024minisplatting} focus on optimizing 3DGS by introducing memory-efficient representation and lightweight pipeline, while Stop-the-Pop~\cite{radl2024stopthepop} improves rendering realism and robustness by a novel hierarchical rasterization approach. Beyond these, Deformable 3DGaussians~\cite{yang2023deformable3dgs} extends 3DGS to efficiently reconstruct and render dynamic scenes by learning a canonical static scene and a time-dependent deformation field for Gaussians. However, both categories are fundamentally data-hungry and show performance degradation under sparse input settings, where their reliance on dense photometric supervision becomes a limiting factor.

\textbf{Novel View Synthesis with sparse views.}
To address the challenges of sparse views, existing methods based on NeRF are mainly divided into two categories: (1) regularization-based methods, which apply techniques such as depth smoothness~\cite{ niemeyer2022regnerf}, occlusion regularization~\cite{yang2023freenerf}, or frequency control to existing sparse data to prevent overfitting; (2) methods incorporating external priors like using pre-trained models to generate depth maps~\cite{chen2021mvsnerf, fan2024instantsplat, jang2021codenerf, yu2021pixelnerf} or feature extractors~\cite{jain2021putting, kwak2023geconerf} to enhance geometric and visual consistency. For 3DGS specifically, some methods have been proposed, such as DNGaussian~\cite{li2024dngaussian} introducing depth regularization, SparseGS~\cite{xiong2024sparsegs} combining depth and diffusion regularization, FSGS~\cite{zhu2024fsgs} increasing point cloud density through Gaussian unpooling, and CoR-GS~\cite{zhang2024cor} and CoherentGS~\cite{paliwal2024coherentgs} utilizing multi-view consistency or monocular depth for optimization. However, existing sparse-view 3D reconstruction methods generally still suffer from severe overfitting, indicating a need for further research to improve their robustness in practical applications.


\section{Method}

Existing Gaussian Splatting methods often suffer from overfitting caused by irregular Gaussian distributions under sparse-view conditions.
To mitigate this limitation, we propose HeroGS, a unified multi-level (image, feature and parameter) guidance framework that provides diverse and complementary supervisory signals to refine and regularize the Gaussian field throughout training.
At the \textbf{image level}, HeroGS introduces pseudo supervision, offering comprehensive guidance for Gaussians across different regions. This enhanced supervision enriches gradient propagation, leading to more globally consistent and well-structured Gaussian field, with camera extrinsics initialized through interpolation and jointly optimized during training.
Building upon this, at the \textbf{feature level}, Feature-Adaptive Densification and Pruning (FADP)  leverages edge-aware and patch-based features extracted from training views to reinforce high-frequency structures while removing redundant Gaussians, resulting in a more accurate and compact representation.
Finally, at the \textbf{parameter level}, Co-Pruned Geometry Consistency (CPG) prunes Gaussians with large parameter discrepancies, effectively suppressing geometric distortions and enhancing global consistency across the scene.
This three-level supervision forms a coherent framework (illustrated in Fig.~\ref{fig:framework}) that not only enhances supervision quality but also preserves structural fidelity under sparse-view conditions.
\begin{figure}[t]
  \centering
  \includegraphics[width=0.95\columnwidth]{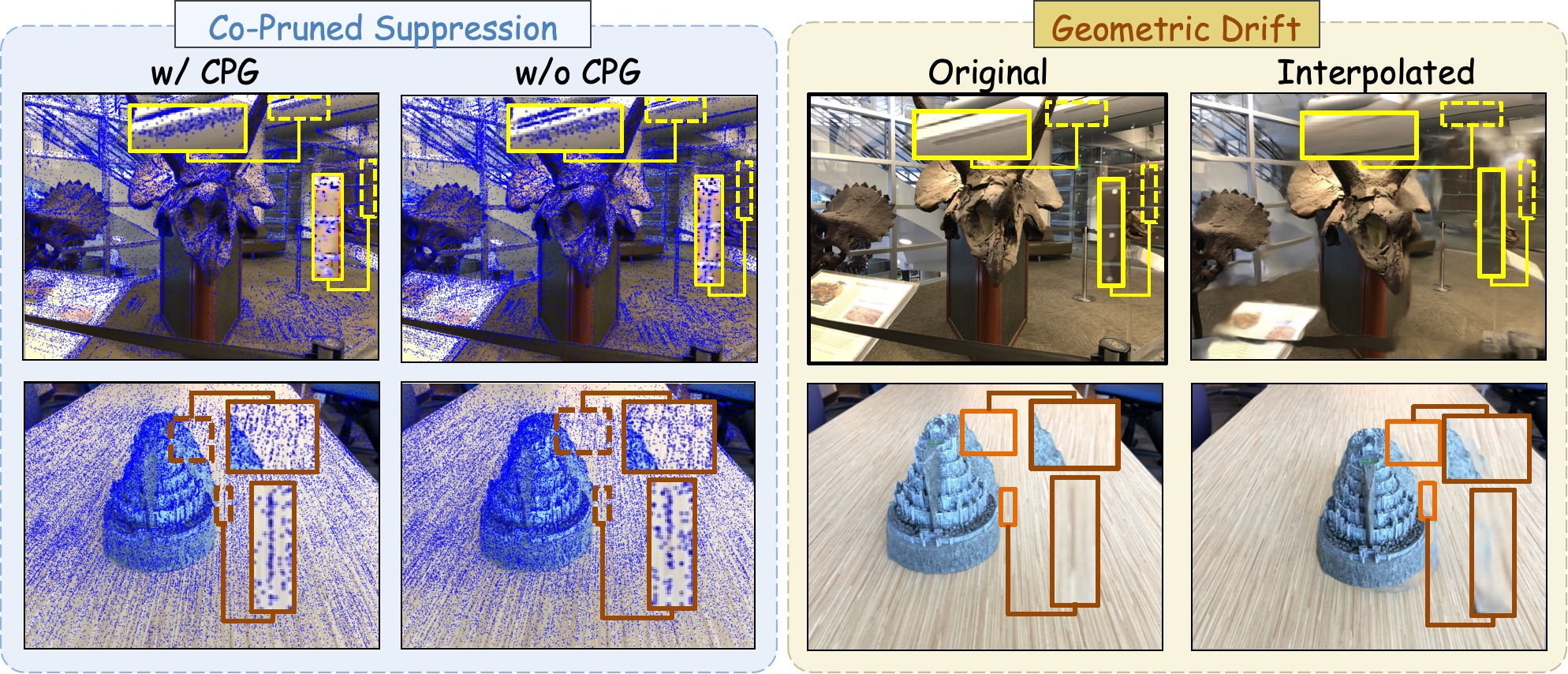}
  \caption{ \textbf{Removing inconsistent Gaussians.} 
%
Columns 3 and 4: original training views and frame-interpolated pseudo-labels, respectively. Although pseudo-labels provide overall supervisory signals, they may lack accuracy in fine details and fail to offer precise guidance for high-frequency geometric structures. Columns 1 and 2: novel-view renderings produced by full pipeline versus those without CPG. It suppresses drifts, yielding sharper edges, coherent surfaces and markedly improved spatial fidelity. 
 }
   \vspace{-11pt}
  \label{fig:VIS_LLFF_geo_inconsist}
\end{figure}
\subsection{Image Level}

Motivated by the observation that denser inputs yield superior reconstructions, we propose to generate intermediate viewpoints between adjacent training views and synthesize their corresponding RGB images as pseudo-labels using a frame interpolation model. Two consecutive training images are defined as $\mathbf{I}_{n}$ and $\mathbf{I}_{n+1}$, with corresponding camera poses $\mathbf{P}_{n}=(\mathbf{R}_{n},\mathbf{T}_{n})$ and $\mathbf{P}_{n+1}=(\mathbf{R}_{n+1},\mathbf{T}_{n+1})$. 
To simplify the notation, we assume a single intermediate frame, denoted as $\mathbf{I}^{(\alpha)}_{n}$, is generated between consecutive training views $\mathbf{I}_{n}$ and $\mathbf{I}_{n+1}$, where $\alpha \in (0,1)$ is the interpolation weight. In our experiments, however, multiple intermediate frames are generated (see Sec.~\ref{sec:analysis} for details).
In this paper, VFI~\cite{seo2025bim}, a state-of-the-art frame interpolation model, is utilized to generate the intermediate images, \emph{i.e.},
\begin{equation}
\small
\mathbf{I}^{(\alpha)}_{n} = \mathrm{VFI}\big(\mathbf{I}_{n}, \mathbf{I}_{n+1},\alpha \big).
\end{equation}Given that our pseudo-labeled images lack ground-truth extrinsic parameters, we initialize the camera pose for each generated image via interpolation. Specifically, spherical linear interpolation (slerp) is employed for rotation and linear interpolation for translation:
\begin{equation}
\small
\begin{split}
\mathbf{R}^{(\alpha)}_n & = \mathrm{slerp}(\mathbf{R}_{n}, \mathbf{R}_{n+1}, \alpha), \\
\mathbf{T}^{(\alpha)}_n & = (1 - \alpha)  \mathbf{T}_{n} + \alpha \mathbf{T}_{n+1}.
\end{split}
\label{Eq(alpha)}
\end{equation}

In the training phase, the scene is rendered from the generated viewpoints, and supervision is applied by comparing the results with the pseudo-labeled images. Let $\hat{\mathbf{I}}_{n}^{(\alpha)}$ be the rendered image from the $n$-th generated viewpoint. The estimated depth of the generated image $\mathbf{I}_{n}^{(\alpha)}$ is denoted as $\mathbf{D}_{n}^{(\alpha)}$, and the rendered depth map from the same viewpoint is $\hat{\mathbf{D}}_{n}^{(\alpha)}$, which is calculated via volumetric rendering:
\begin{equation}
\small
\hat{\mathbf{D}}_{n}^{(\alpha)} = \sum_{i=1}^{L} d_i o_i \prod_{j=1}^{i-1}(1-o_j).
\end{equation}The sum is over the $L$ ordered 3D Gaussians that overlap a pixel. For the $i$-th Gaussian, ${d}_{i}$ is its depth and $o_{i}$ is its learned opacity.
Similar to \cite{zhu2024fsgs}, our training objective integrates photometric supervision on the generated images, measured by L1 and D-SSIM losses, with geometric supervision on the rendered depth maps, assessed using a Pearson correlation coefficient loss, \emph{i.e.},
\begin{align}
\footnotesize
    \mathcal{L}_g = \sum_{n=1}^{N-1} \bigg( & \lambda_1 \|\mathbf{I}_{n}^{(\alpha)} - \hat{\mathbf{I}}_{n}^{(\alpha)}\|_1 + \lambda_2 \mathcal{L}_{\text{D-SSIM}}(\mathbf{I}_{n}^{(\alpha)}, \hat{\mathbf{I}}_{n}^{(\alpha)}) \notag \\
    & + \lambda_3 \big(1 - \mathrm{Corr}(\mathbf{D}_{n}^{(\alpha)}, \hat{\mathbf{D}}_{n}^{(\alpha)})\big) \bigg).
\label{equa(3)}
\end{align}

To ensure reliable supervision, low-quality pseudo-labels are filtered out by a selection module, and only the high-quality ones are retained for training. Further details are discussed in appendix.

Since the frame interpolation model lacks 3D awareness, a slight mismatch can occur between generated images and their camera poses. To mitigate this, these interpolated camera poses are parameterized as learnable variables, optimizing them jointly with the Gaussian field throughout training.
\subsection{Feature Level}

While pseudo-labels offer valuable global supervision, their limited precision in fine details makes it difficult to accurately constrain high-frequency geometric structures (Fig.~\ref{fig:VIS_LLFF_geo_inconsist}), motivating further refinement at feature level.

To resolve this concern and enhance geometric detail, \textbf{F}eature-\textbf{A}daptive \textbf{D}ensification and \textbf{P}runing (\textbf{FADP}) is introduced. FADP refines Gaussian distributions in high-frequency regions by improving texture-aware geometry through two complementary strategies: edge-aware densification and patch-based density control.

For each training image $\mathbf{I}_{n}$, we first extract its corresponding edge map $\mathbf{E}_{n}$ using an edge detection model~\cite{liufu2025sauge}. 2D points are then sampled along the detected edges and back-projected into 3D space to define the centers of new Gaussians. The attributes of each new Gaussian $\hat{\mathbf{G}}$, namely color, opacity, and shape, are initialized based on its neighbors. Specifically, we identify its $K$-nearest neighbors ($K=3$ by default), $\{\hat{\mathbf{G}}_{1}, \hat{\mathbf{G}}_{2},\dots, \hat{\mathbf{G}}_{K}\}$, from the existing Gaussians. Each attribute $\hat{\mathbf{A}}$ of the new Gaussian $\hat{\mathbf{G}}$ is computed via inverse-distance weighted interpolation:
\begin{equation}
\small
\hat{\mathbf{A}} = \frac{\sum_{k=1}^{K} w_{k} \cdot \mathbf{A}_{k}}{\sum_{k=1}^{K} w_{k}}, \quad w_{k} = \frac{1}{d_{k} + \epsilon},
\end{equation}
where $\mathbf{A}_{k}$ is the attribute of the $k$-th nearest inherent $\mathbf{G}_{k}$, $d_{k}$ is its Euclidean distance to $\hat{\mathbf{G}}$, and $\epsilon$ is a small constant. 

To complement edge-aware densification, a \textbf{patch-based density controlling strategy} is proposed to prevent oversampling and promote balanced Gaussian distributions. Specifically, each training image is divided into an $m \times m$ grid of patches ($m=8$ by default) and Gaussians are projected onto image plane. Let $\mathcal{C} = \{ c_1, c_2, \ldots, c_{m^2} \} \in \mathbb{R}^{m^2}$ denote the number of projected Gaussians in each of the $m \times m$ patches. We reweight these counts to obtain $\mathcal{C'} = \{ c_1', c_2', \ldots, c_{m^2}' \}$ using the following piecewise function:
\begin{equation}
\small
c_i' =
\begin{cases}
c_{\mathrm{min}}, & \text{if } c \leq \tau_{\mathrm{sparse}} \\
c_i \cdot \lambda_{\mathrm{low}}, & \text{if } \tau_{\mathrm{sparse}} < c < \tau_{\mathrm{low}} \\
c_i, & \text{if } \tau_{\mathrm{low}} \leq c \leq \tau_{\mathrm{high}} \\
c_i \cdot \lambda_{\mathrm{high}}, & \text{if } c > \tau_{\mathrm{high}}
\end{cases}
\end{equation}
where $\tau_{\mathrm{sparse}}, \tau_{\mathrm{low}}, \tau_{\mathrm{high}}$ are density thresholds, and $\lambda_{\mathrm{low}} > 1$, $\lambda_{\mathrm{high}} < 1$ are scaling factors that increase sampling in underrepresented regions and suppress sampling in over-dense areas. A minimum count $c_{\mathrm{min}}$ is enforced in sparse patches to guarantee coverage.

To maintain the global number of Gaussians, we apply normalization:
\begin{equation}
\small
\mathcal{C'} \leftarrow \text{round}\left( \mathcal{C'} \cdot \frac{\sum_i c_i}{\sum_i c_i'} \right).
\end{equation}

A residual correction step is applied to ensure $\sum_i c_i' = \sum_i c_i$ by adjusting the entries in $\mathbf{c'}$.

In essence, the globally refined Gaussian distributions obtained from the previous stage provide a more complete structural foundation, bringing $\mathcal{C}$ closer to the distributions required for realistic rendering even before optimization. This enables patch-based
density controlling strategy to achieve more effective and reliable refinement.

By jointly leveraging edge-based guidance and patch-based controlling, FADP achieves an optimal balance between texture-sensitive densification and globally consistent Gaussian distributions. The two strategies are tightly coupled: while edge-based densification focuses on adding high-frequency details near object boundaries, patch-based controlling ensures that such additions do not result in local over-concentration or sparsity elsewhere. This synergy enables effective geometry refinement without disrupting the global spatial coherence of the scene.
\begin{table*}[htp!]

  \begin{tabularx}{\textwidth}{l|*{9}{Y}}

  \hline
  \multirow{2}{*}{Methods} & \multicolumn{3}{c}{2 views} & \multicolumn{3}{c}{3 views} & \multicolumn{3}{c}{6 views}  \\
  & PSNR↑ & SSIM↑ & LPIPS↓ & PSNR↑ & SSIM↑ & LPIPS↓ & PSNR↑ & SSIM↑ & LPIPS↓ \\
  \hline
  Mip-NeRF~\cite{barron2021mipnerf}      & - & - & - & 16.11 & 0.401 & 0.460 & 22.91 & 0.756 & 0.213\\
  SparseNeRF~\cite{wang2023sparsenerf}   & 17.80 & 0.532 & 0.372 & 19.86 & 0.624 & 0.328 & 23.26 & 0.741 & 0.235\\
  FrugalNeRF~\cite{Lin_2025_CVPR} & \underline{18.29} & \underline{0.557} & 0.342 & 19.92 & 0.634 & 0.297 & - & - & -  \\
  \hline
  3DGS~\cite{kerbl20233d}     & 16.19 & 0.437 & 0.388 & 19.24 & 0.649 & 0.237 & 23.63 & 0.809 & 0.129\\
  DRGS~\cite{chung2023depth}      & 17.04 & 0.513 & \underline{0.318} & 16.73 & 0.487 & 0.310 & 18.60 & 0.560 & 0.239\\
  DNGaussian~\cite{li2024dngaussian}   & 17.03 & 0.519 & 0.362 & 19.12 & 0.591 & 0.294 & 22.18 & 0.755 & 0.198\\
  FSGS~\cite{zhu2024fsgs}   & 15.65 & 0.460 & 0.412 & 20.43 & 0.682 & 0.248 & 24.15 & 0.823 & \underline{0.128}\\
  CoR-GS~\cite{zhang2024cor}      & 17.38 & 0.539 & 0.350 & 20.45 & \underline{0.712} & \underline{0.196} & 24.29 & 0.824 & 0.135 \\
  
  DropGaussian~\cite{Park_2025_CVPR}      & 17.32 & 0.509 & 0.343 & \underline{20.55} & 0.710 & 0.200 & \underline{24.55} & \underline{0.835} & \textbf{0.117} \\
  \hline
  HeroGS (Ours)  & \textbf{18.78} & \textbf{0.595} & \textbf{0.317} & \textbf{21.30} & \textbf{0.739} & \textbf{0.189} & \textbf{24.59} & \textbf{0.837} & 0.135 \\
  
  \hline

  \end{tabularx}
  \caption{\textbf{Quantitative results on LLFF with 2, 3, 6 training views.} The \textbf{best} and \underline{second-best} entries are denoted using \textbf{bold} and \underline{underline} respectively. 3DGS-based methods require multi-view stereo estimation from COLMAP, which fails on 3 scenes for 2 training views. We therefore report the averaged metrics of the remaining scenes.}
  \label{tab:QUA_LLFF}
\end{table*}
\begin{table*}[htp!]

  \begin{tabularx}{\textwidth}{l|*{9}{Y}}

  \hline
  \multirow{2}{*}{Methods} & \multicolumn{3}{c}{2 views} & \multicolumn{3}{c}{3 views} & \multicolumn{3}{c}{6 views}  \\
  & PSNR↑ & SSIM↑ & LPIPS↓ & PSNR↑ & SSIM↑ & LPIPS↓ & PSNR↑ & SSIM↑ & LPIPS↓ \\
  \hline
  3DGS      & 16.19 & 0.437 & 0.388 & 19.24 & 0.649 & 0.237 & 23.63 & 0.809 & \underline{0.129}\\
  FSGS   & 15.65 & 0.460 & 0.412 & 20.43 & 0.682 & 0.248 & 24.15 & 0.823 & \textbf{0.128}\\
  \hline
  +VFI     & 16.91 & 0.489 & 0.384 & 20.68 & 0.704 & 0.204 & 24.18 & 0.822 & 0.135\\
  +FADP      & 17.28 & 0.503 & 0.370 & 20.99 & 0.720 & 0.192 & 24.25 & 0.824 & 0.136\\
  +GSField (freeze$_{scale}$)   & \underline{17.93} & \underline{0.564} & \underline{0.345} & \underline{21.08} & \underline{0.732} & \underline{0.191} & \underline{24.40} & \underline{0.832} & 0.136\\

  HeroGS (Ours)  & \textbf{18.78} & \textbf{0.595} & \textbf{0.317} & \textbf{21.30} & \textbf{0.739} & \textbf{0.189} & \textbf{24.59} & \textbf{0.837} & 0.135 \\
  
  \hline
  \end{tabularx}
  \caption{Ablation study on LLFF dataset with 2, 3, 6 training views. The \textbf{best} and \underline{second-best} entries are denoted using \textbf{bold} and \underline{underline} respectively. 
  Under the 2-view setting, the baseline method (FSGS~\cite{zhu2024fsgs}) performs poorly, while our approach achieves a significant performance improvement.
  }  
  \label{tab:Ablation_LLFF}
\end{table*}

\subsection{Parameter Level}

Motivated by CoR-GS~\cite{zhang2024cor} and aimed at eliminating erroneous Gaussians, we introduce the \textbf{C}o-\textbf{P}runed \textbf{G}eometry Consistency (\textbf{CPG})  at parameter level. CPG incorporates two auxiliary Gaussian fields, which are trained jointly with the primary field. To facilitate more effective co-pruning, the parameters of the auxiliary Gaussian fields are frozen after a predefined training iteration. By comparing the consistency of corresponding Gaussians across all three fields, we identify and remove inconsistent points through a comprehensive co-pruning strategy. This filtering process effectively suppresses geometric artifacts, such as blurriness and shape distortion, leading to significantly improved spatial coherence, as illustrated in Fig.~\ref{fig:VIS_LLFF_geo_inconsist}.

\textbf{Training Strategy.} We adopt a two-stage strategy. Before a predefined iteration number \( N_{\mathrm{iter}} \), all three Gaussian Splatting (GS) fields perform \emph{mutual co-pruning}. Subsequently, the two auxiliary fields are partially frozen (fixing scale and rotation), and only the primary GS field continues to be updated. The pruning becomes unilateral: the primary field is pruned based on geometric agreement with the two frozen auxiliary fields.

\textbf{Co-Pruning Criterion.} Given a source GS field \( \mathcal{G}_s = \{ \mathbf{G}^s_1, \dots, \mathbf{G}^s_{Y} \} \) and a target field \( \mathcal{G}_t \), let \( \mathbf{p}_y^s, \mathbf{p}_z^t \in \mathbb{R}^3 \) denote the 3D position of \( \mathbf{G}^s_y \) and \( \mathbf{G}^t_z \), respectively. For each Gaussian in $\mathcal{G}_s$, its nearest neighbor in the target field is obtained as follows:
\begin{equation}
\small
z^* = \arg\min_z \| \mathbf{p}_y^s - \mathbf{p}_z^t \|_2,
\end{equation}

\begin{equation}
\small
w_y = \| \mathbf{p}_y^s - \mathbf{p}_{z^*}^t \|_2.
\end{equation}

\( \mathbf{G}^s_y \) is pruned from source field if the distance exceeds a threshold $w_y > \delta$, where we set the threshold $\delta=5$. 
Guided by parameter level, Gaussian fields contain finer details (as highlighted by the orange points in Fig.~\ref{fig:framework}), making it easier to accurately identify erroneous Gaussian \( \mathbf{G}^s_y \).

\textbf{Post-Freeze Behavior.} After \( N_{\mathrm{iter}} \), pruning is applied solely to the primary field, leveraging both auxiliary fields as geometric references. This strategy facilitates progressive refinement through early-stage alignment via multi-field redundancy and late-stage geometry stabilization.
Throughout training, each GS field, $i.e$., the primary and the two auxiliary fields—is independently supervised using a combination of training view and generated view losses:
\begin{equation}
\small
\mathcal{L} = \lambda_{g}\mathcal{L}_{g}+\mathcal{L}_{r},
\end{equation}
where $\mathcal{L}_{g}$ and $\mathcal{L}_{r}$ follow the same formulation as in Eq.~\ref{equa(3)}.

\section{Experiments}
\subsection{Setup}
\begin{table}[t]
\centering
\renewcommand{\arraystretch}{1.1}
\setlength{\tabcolsep}{1pt}
\resizebox{\columnwidth}{!}{
\begin{tabular}{l|ccc ccc}
\hline
\multirow{2}{*}{\textbf{Methods}} 
& \multicolumn{3}{c}{\textbf{3 Views}} 
& \multicolumn{3}{c}{\textbf{6 Views}} \\ 
& PSNR$\uparrow$ & SSIM$\uparrow$ & LPIPS$\downarrow$ 
& PSNR$\uparrow$ & SSIM$\uparrow$ & LPIPS$\downarrow$ \\ 
\hline
3DGS      & 16.39 & 0.442 & \underline{0.432} & 22.49 & 0.725 & 0.279 \\
FSGS      & 16.88 & 0.474 & 0.434 & 23.63 & 0.742 & 0.281 \\
CoR-GS   & \underline{17.06} & \underline{0.491} & 0.443 & 23.59 & 0.742 & 0.275 \\
DropGaussian & 16.81 & 0.480 & 0.438 & \underline{24.15} & \underline{0.772} & \textbf{0.215} \\
\textbf{HeroGS}      & \textbf{17.51} & \textbf{0.512} & \textbf{0.422} & \textbf{24.70} & \textbf{0.781} & \underline{0.272} \\
\hline
\end{tabular}
}

\caption{\textbf{Quantitative results on Tanks\&Temples with 3, 6 training views without downsampling.}}
\label{tab:QUA_T&T}
\end{table}
HeroGS is implemented based upon FSGS~\cite{zhu2024fsgs}, and is evaluated on the LLFF~\cite{mildenhall2019llff} and Tanks\&Temples~\cite{knapitsch2017tanks} datasets. On LLFF, we use 2, 3, and 6 training views, with resolutions downsampled by 8$\times$. Settings for 3 and 6 training views follow previous work~\cite{jang2025comapgs}. For the 2-view case, as FSGS~\cite{zhu2024fsgs} requires COLMAP-based multi-view stereo which fails on 3 scenes, we evaluate on the remaining 5 scenes. Here, two views are randomly chosen for training, and the rest for testing. On Tanks\&Temples, 3 and 6 training views are used without downsampling. In our experiments, a subset of training views is uniformly sampled, with remaining views used for testing. Following prior works~\cite{zhu2024fsgs,zhang2024cor}, PSNR, SSIM, and LPIPS are employed as the evaluation metrics.

\subsection{Comparison}
\begin{figure*}[t]
  \centering
  \includegraphics[width=0.85\textwidth]{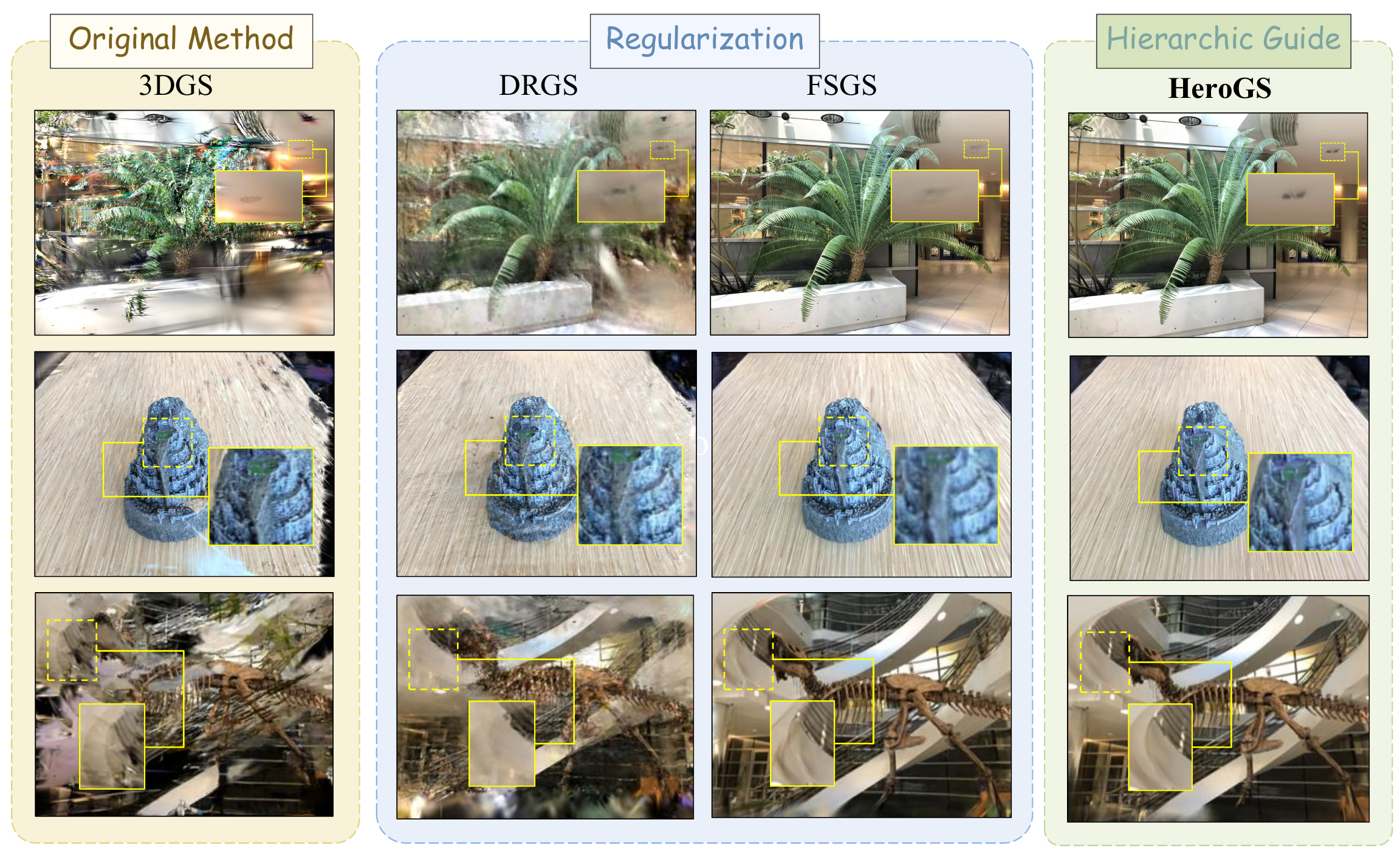}
  \caption{ \textbf{Qualitative Comparison on LLFF (3 training views).} Under extreme view sparsity, 3DGS~\cite{kerbl20233d} collapses into severe artifacts and blurred geometry. DRGS and FSGS recover coarse structure yet still exhibit over-smoothed textures and noisy backgrounds. In contrast, HeroGS guides the model from complementary levels to deliver markedly sharper object boundaries, richer high-frequency textures, and distinctly clearer distant regions—demonstrating the efficacy of our full pipeline.
 }
  \vspace{-5pt}
  \label{fig:LLFF_vis}
\end{figure*}

\textbf{LLFF.}
Quantitative results are presented in Tab.~\ref{tab:QUA_LLFF}, and qualitative comparisons are shown in Fig.~\ref{fig:LLFF_vis}. HeroGS achieves superior performance across multiple metrics. Notably, under the challenging 2-view training setting, HeroGS demonstrates a significant performance advantage over existing baselines. This substantial gain is primarily driven by the strategic introduction of hierarchical guidance, effectively compensating for the scarcity of reliable supervision in extremely sparse input views.

Moreover, as Fig.~\ref{fig:LLFF_vis} illustrates, 3DGS struggles to recover the structure of some objects, while DRGS tends to synthesize smooth views lacking high-frequency details, compared to FSGS and HeroGS. Although FSGS recovers most structural details, it often fails to produce accurate local texture patterns. In contrast, HeroGS renders more accurate and detailed textures, visible in the intricate patterns of fortress and trex, and produces clearer background regions, such as the distant structures in fern. This visual fidelity largely stems from our paradigm’s ability to mitigate overfitting by guiding Gaussian distributions at multiple levels.


\textbf{Tanks\&Temples.}
Extensive evaluations on the Tanks\&Temples dataset are conducted to assess the effectiveness of HeroGS in complex and large-scale environments. As shown in Fig.~\ref{fig:T&T_vis} (Appendix) and Tab.~\ref{tab:QUA_T&T}, HeroGS is compared against several 3DGS-based baselines.
The basic 3DGS struggles to preserve geometric and textural fidelity under limited supervision, while prior approaches~\cite{zhu2024fsgs,zhang2024cor} introduce sparse or noisy regularization, often resulting in oversmoothed geometry and noticeable artifacts.
DropGaussian~\cite{Park_2025_CVPR} further exhibits ghosting effects due to inaccurate Gaussian distributions in high-frequency regions.
In contrast, the proposed HeroGS framework injects pseudo dense supervision at the image level, effectively bridging the sparse-to-dense gap.
Furthermore, FADP and CPG collaboratively refine Gaussian distributions, yielding more accurate reconstructions of fine-grained structures and high-frequency textures.
Notably, HeroGS consistently achieves superior PSNR and SSIM scores, demonstrating robust generalization and stability under sparse-input conditions.
\subsection{Ablation study}
\begin{figure}[h]
  \centering
  \includegraphics[width=1.0\columnwidth]{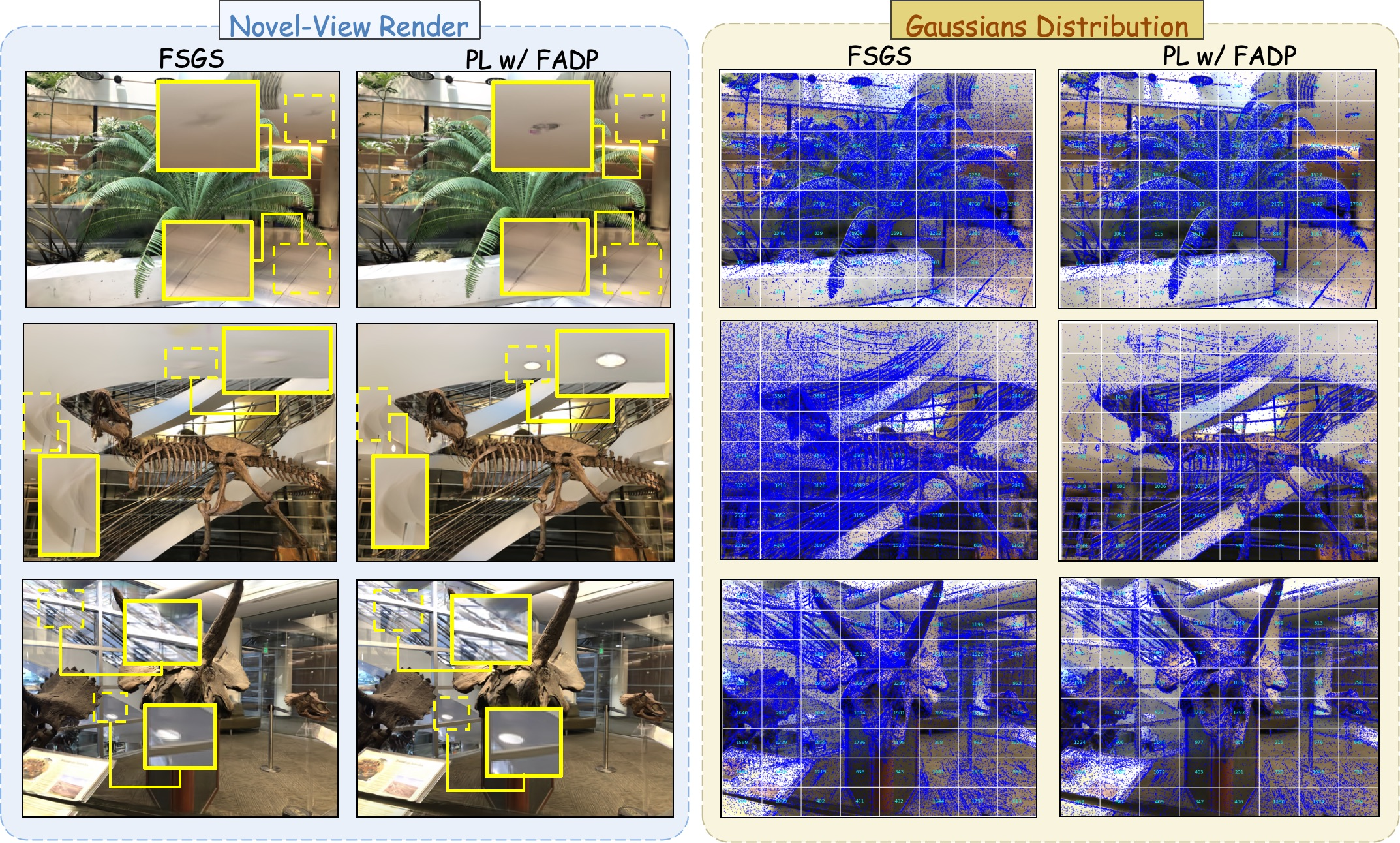}
  \caption{ \textbf{Comparison with and without image and feature level guidance.} Columns 1 and 2: novel-view renderings produced by baseline versus baseline with pseudo-labels and FADP. Columns 3 and 4: their corresponding  Gaussians. 
 }
  \vspace{-10pt}
  \label{fig:FADP&PL_compare}
\end{figure}
We conduct comprehensive ablation studies on the LLFF dataset, encompassing both qualitative and quantitative comparisons. As shown in Tab.~\ref{tab:Ablation_LLFF}, simply incorporating image-level guidance significantly improves performance over the baseline (FSGS), especially in extremely sparse view settings. This highlights the effectiveness of using interpolated frames as pseudo-supervision to guide the model.
Additionally, Fig.~\ref{fig:FADP&PL_compare} demonstrates that integrating the FADP into the proposed framework results in clearer reconstructions and more reasonable Gaussian distributions. Notably, while the total number of Gaussians is reduced compared to the baseline, their density increases near object boundaries. This contributes to improved geometric fidelity and the preservation of high-frequency details.
To validate the effectiveness of multiple GS fields, an ablation study is conducted on their number and configuration. HeroGS employs two additional GS fields, with scale and rotation frozen respectively. 
This setup is compared against: 
(1) using a single additional GS field with its scale frozen, and (2) the baseline with only one GS field. As Tab.~\ref{tab:Ablation_LLFF} (in the Appendix) indicates, HeroGS achieves superior performance, demonstrating that disentangled geometric representation improves reconstruction accuracy and detail preservation. Fig.~\ref{fig:VIS_LLFF_geo_inconsist} provides further qualitative comparison. We observe that the interpolation network alone fails to accurately recover fine details and textures on interpolated images (Column 4) when compared with the training view (Column 3). When CPG is disabled (Column 2), the rendering suffers from slight geometric misalignment of textures. In contrast, enabling CPG (Column 1) in our paradigm effectively ameliorates both geometric drift and texture degradation, as it prunes Gaussians with spatial deviations and preserves those aligned with consistent geometry.

\subsection{Analysis}
\label{sec:analysis}
\begin{table}[t]
\centering

  \begin{tabularx}{0.5\textwidth}{l|*{3}{Y}}

  \hline
  Settings & PSNR & SSIM & LPIPS\\
  \hline
  w/o Post-Freeze      & 21.16 & 0.735 &  0.190\\
  All     & \textbf{21.30} & \textbf{0.739} & \textbf{0.189}\\
  \hline
  
  \end{tabularx}
  \caption{Ablation study on Post-Freeze Behavior on LLFF dataset for 3 training views. }
  \label{tab:Ablation_post_freeze}
\end{table}
\textbf{Impact of Post-Freeze Behavior.}
To illustrate the effect of the parameter-freezing strategy, the Post-Freeze Behavior is ablated in Tab.~\ref{tab:Ablation_post_freeze}. The frozen parameters compel the two auxiliary Gaussian fields to learn distinct local optima. This, in turn, facilitates the primary Gaussian field in effectively pruning misplaced or ill-shaped Gaussians, thereby yielding more authentic texture details.

\begin{figure}[t]
  \centering
  \includegraphics[width=0.95\columnwidth]{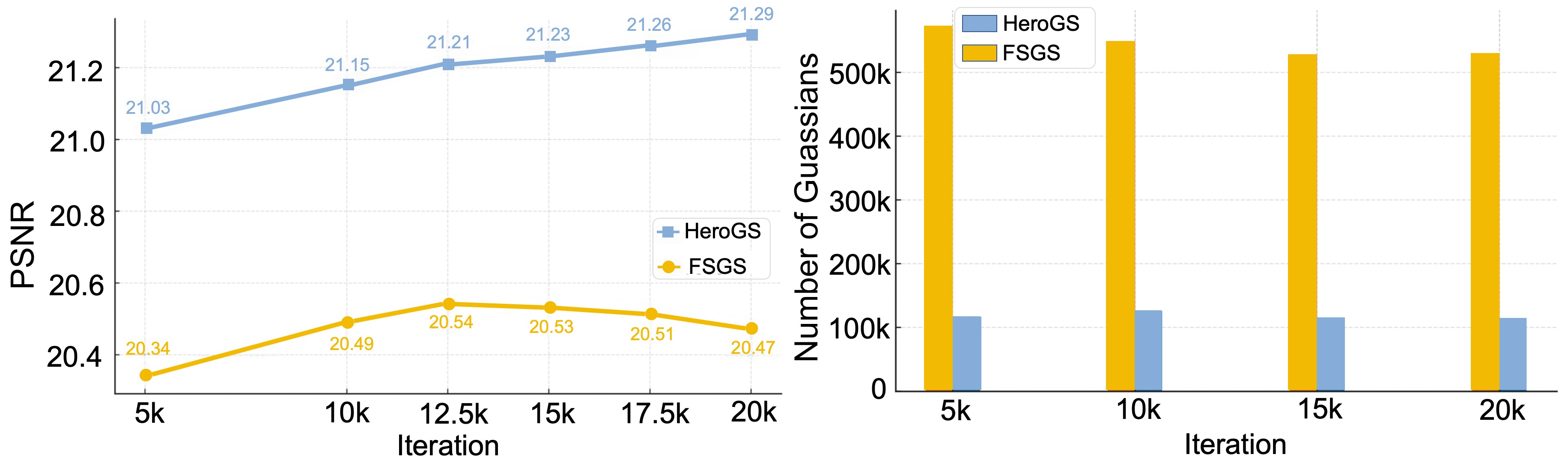}
  \caption{ 
  \textbf{Analysis of training dynamics.}
  Comparison of PSNR (left) and the number of Gaussians (right) between our HeroGS and baseline during training.
 }
  \label{fig:training}
\end{figure}

\textbf{Analysis of Training Dynamics.}
We further analyze the training dynamics by monitoring the variation of Gaussian counts and PSNR values throughout the optimization process on the LLFF dataset, as illustrated in Fig.~\ref{fig:training}. As training progresses, our method exhibits a steady improvement in PSNR, indicating consistent convergence and stable optimization, whereas the baseline shows noticeable fluctuations and even performance degradation in later stages. Notably, at around 5K iterations, our approach already surpasses the baseline by a clear margin, while maintaining a significantly lower number of Gaussians. This reduction demonstrates that HeroGS achieves more compact scene representation, leading to improved memory efficiency and faster rendering speed.

\textbf{Interpolation Factor.}
The interpolation factor, denoted as $S$, represents the number of frames generated for supervision between two consecutive training views. Specifically, for Eq.~\ref{Eq(alpha)}, $S-1$ views are generated between the $n$-th and $(n+1)$-th frames, in which case $\alpha$ takes values from the set $\{ \frac{1}{S}, \frac{2}{S}, \dots, \frac{S-1}{S} \}$, leading to $S-1$ intermediate camera extrinsics evenly distributed along the motion path.
As Fig.~\ref{fig:Inter_factor} (in the Appendix) shows, performance metrics generally exhibit lower values at an interpolation factor of $S=2$. While performance improves significantly from $S=4$ onwards, including at $S=8$ and $S=16$, all metrics demonstrate a tendency to stabilize beyond $S=4$. An interpolation factor of $S=4$ is selected in the experiments to strike an optimal balance between visual quality and computational cost.

\textbf{Frame Interpolation Model Comparisons.}
Tab.~\ref{tab:Comp_inter_model} presents a comparative analysis of the effects of employing different frame interpolation models to generate pseudo-labels. The first row showcases the performance when using alternative frame interpolation model~\cite{Wu_2024_CVPR}. The second row demonstrates the results achieved with our chosen model. The third row provides the outcome when the interpolated images are replaced by uniformly sampled ground truth images, while keeping all other weights constant. To clearly demonstrate the efficacy of parameter level, CPG is excluded in all experiments above. As shown in the table, utilizing the interpolated results from our chosen frame interpolation model at image level yields superior performance compared to other models. However, a minor gap still persists when compared to using ground truth images. This discrepancy arises from the presence of irregularly distributed Gaussians, which leads to local geometric misalignment and texture inconsistencies. CPG effectively mitigates the impact of these inconsistencies, thereby enhancing model performance and bridging this gap.

\begin{table}[h]
\centering

  \begin{tabularx}{0.45\textwidth}{l|*{3}{Y}}

  \hline
  Settings & PSNR & SSIM & LPIPS\\
  \hline
  PerVFI~\cite{Wu_2024_CVPR}      & 20.78 & 0.711 &  0.204\\
  BIMVFI     & 20.99 & 0.720 & 0.192\\
  Ground Truth     & 21.09 & 0.720 & 0.195\\
  Ours & \textbf{21.30} & \textbf{0.739} & \textbf{0.189}\\
  \hline

  \end{tabularx}
  \caption{Comparison between different Interpolation Model on LLFF dataset for 3 training views. }  
  \label{tab:Comp_inter_model}
\end{table}

\begin{table}[t]
\centering
\setlength{\tabcolsep}{4pt}
\resizebox{\columnwidth}{!}{
\begin{tabularx}{\columnwidth}{cccc|ccc}

\toprule
   &\textbf{PL} & \textbf{FADP} & \textbf{CPG} & \textbf{PSNR} $\uparrow$ & \textbf{SSIM} $\uparrow$& \textbf{LPIPS} $\downarrow$ \\
\midrule \multirow{4}{*}{\shortstack{One\\ Level}}& \cmark            &             &               & 16.91 & 0.489&0.384 \\  &          &    \cmark         &               & 16.64 & 0.483&0.390 \\   &         &             &    \cmark           & 16.78  & 0.535 &0.405
\\  &\multicolumn{3}{c|}{\textbf{average}}&16.78&0.502&0.393 \\\hline \multirow{4}{*}{\shortstack{Two\\Levels}} &\cmark            & \cmark             &               & 17.28 & 0.503&0.370 \\ &
 \cmark             &               & \cmark             & 18.33 & 0.577& 0.336 \\ &
             &     \cmark          & \cmark             & 17.43 & 0.540& 0.372\\ 
 & \multicolumn{3}{c|}{\textbf{ average}}&17.68&0.540&0.359 \\ \hline &
\cmark             & \cmark             & \cmark             & \textbf{18.78} & \textbf{0.595}&\textbf{0.317} \\
\bottomrule
\end{tabularx}}
\caption{\textbf{Ablation study on level interdependencies} on LLFF dataset for 2 training views.}
\label{tab:ablation2views}
\vspace{-5pt}
\end{table}

\textbf{Level Interdependency.}
The numerical trend in Tab.~\ref{tab:ablation2views} reveals clear hierarchical synergy among the three levels.
On average, adding a second level yields gains of $+0.9$ PSNR, $+0.038$ SSIM, and $-0.034$ LPIPS, indicating complementary interaction.
Integrating all three levels further improves performance by $+1.1$ PSNR, $+0.055$ SSIM, and $-0.042$ LPIPS, surpassing the previous stage.
The difference in incremental gains further indicates the presence of intrinsic coupling across levels, where improvements at one level reinforce and amplify others. This interdependency implies that the modules do not function in isolation but collaboratively contribute to a unified optimization process.

\section{Conclusion}
In this work, we present \textbf{HeroGS}, a tightly coupled hierarchical guidance framework that tames the ill-posed challenge of sparse-view 3D reconstruction.
At the \textit{image level}, pseudo-labels transform sparse supervision into pseudo-dense guidance, providing global regularization and rich structural cues for feature level.
Building upon this, \textbf{FADP} at the \textit{feature level} refines local geometry by injecting texture-aware high-frequency feature cues.
The \textbf{CPG} at the \textit{parameter level} further enforces geometric consistency via self-supervised co-pruning. These three levels form a hierarchical guidance framework that optimizes the overall Gaussian distributions.
Extensive experiments on diverse benchmarks demonstrate that HeroGS sets a new state of the art for sparse-view reconstruction.
\section*{Acknowledgments}
This work was supported in part by the Key Deployment Program of the Chinese Academy of Sciences, China under Grant KGFZD-145-25-39, the National Natural Science
Foundation of China under Grants 62272438, and Beijing
Natural Science Foundation L25700.

{
    \small
    \bibliographystyle{ieeenat_fullname}
    \bibliography{main}
}
\clearpage
\setcounter{page}{1}
\maketitlesupplementary
\section{Implementation Details} 

We conduct training for 20k iterations and set \( N_{\mathrm{iter}}=10k\) for both datasets. Feature Adaptive Densification and Prune is applied at 10k iterations and patch-based density controlling strategy at 8k iterations for stability. Further more, we set \( \lambda_{\text{low}} = 2.0 \) and \( \lambda_{\text{high}} = 0.8 \). In addition, the parameter $\tau_{\text{sparse}}$ is set to be the number of Gaussians in the top 90$\%$ of patches sorted by point density in descending order, while $\tau_{\text{high}}$ to be the number of top 10$\%$. Besides, the number of generated views is set to 4×, meaning that 3 generated views are inserted between every two training views. Frame interpolation is introduced after 2k iterations, with its loss evaluated at an interval of every 10 iterations. Starting from iteration 2000, it is applied for the first 100 iterations of every subsequent 200-iteration cycle, and disabled for the remaining 100 iterations. We first set $\lambda_{g}=0.075$, which increases as the training iterations grows. SAUGE~\cite{liufu2025sauge}, a model based on SAM, is used to extract the edge of each training image. The co-pruning parameters are set following the configuration used in CoR-GS~\cite{zhang2024cor}. HeroGS is initialized with point clouds and precomputed camera poses from COLMAP.

\textbf{Selection Module.}
The Gaussian fields are first trained for 2000 iterations without image-level guidance.
After this stage, pseudo-label images are filtered based on their quality, and the high-quality ones are used for subsequent supervision. 
The selection metric is computed as:
\begin{equation}
M = \lambda_1 \| I^{\alpha} - \hat{I}^{\alpha} \| + \lambda_2 \mathcal{L}_{D\text{-}SSIM} \, \mathrm{Cor}(\hat{I}^{\alpha}, I^{\alpha}),
\end{equation}
where $I^{\alpha}$ denotes a pseudo-label, and $\hat{I}^{\alpha}$ represents the corresponding rendered image. 
To avoid the influence of model instability during the early training phase, we progressively re-evaluate and re-select pseudo-labels according to the rendered outputs as training proceeds.
If there are $N$ pseudo-labeled images in total, $N/2$ images with the smallest values of $M$ (i.e., those closest to the rendered results) are selected as high-quality supervision for the subsequent training phase.

\begin{figure}[h]
  \centering
  \includegraphics[width=0.95\columnwidth]{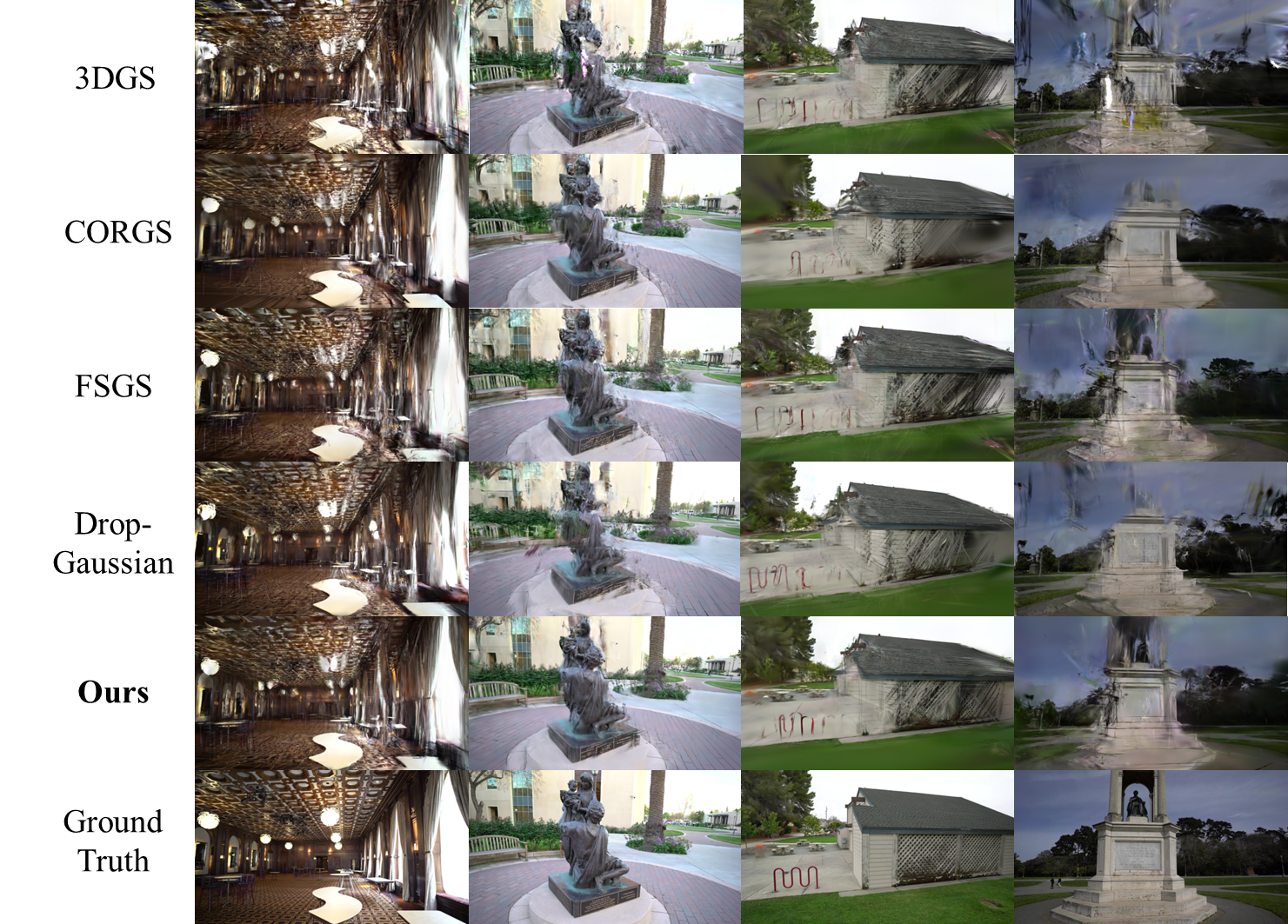}
  \caption{\textbf{Qualitative Comparison on Tanks for 3 training views.} In large-scale dataset, 3DGS and DropGaussian struggles to maintain geometry and texture fidelity, exhibiting significant artifacts. FSGS and CoR-GS recover coarse structure yet still exhibit over-smoothed geometry and artifacts. In contrast, HeroGS reconstruct fine structures and high-frequency textures.
  }
  \label{fig:T_T_vis}
\end{figure}

\section{More Comparison Results}
\begin{table}[t]
\centering
\renewcommand{\arraystretch}{1.1}
\setlength{\tabcolsep}{1pt}

  \begin{tabularx}{\columnwidth}{l|*{6}{Y}}

  \hline
  \multirow{2}{*}{Methods} & \multicolumn{3}{c}{12 views} & \multicolumn{3}{c}{24 views} \\
  & PSNR↑ & SSIM↑ & LPIPS↓ & PSNR↑ & SSIM↑ & LPIPS↓\\
  \hline
  3DGS      & 18.44 & 0.521 & 0.385 & 23.22 & 0.730 & 0.234\\
  FSGS      & 18.93 & 0.539 & 0.380 & 23.46 & 0.738 & 0.237\\
  CoR-GS      & \underline{19.59} & \underline{0.578} & 0.374 & 23.39 & 0.727 & 0.272\\
  DropGaussian      & 19.49 & 0.573 & \textbf{0.366} & \underline{24.03} & \underline{0.762} & \textbf{0.225}\\
  HeroGS      & \textbf{19.99} & \textbf{0.591} & \underline{0.373} & \textbf{24.18} & \textbf{0.766} & \underline{0.229}\\
  \hline

  \end{tabularx}\
  \caption{\textbf{Quantitative results on Mip-NeRF360~\cite{Barron_2022_CVPR} with 12, 24 training views.} }
  \label{tab:QUA_Mip}
\end{table}

\begin{figure*}[h]
  \centering
  \includegraphics[width=0.85\textwidth]{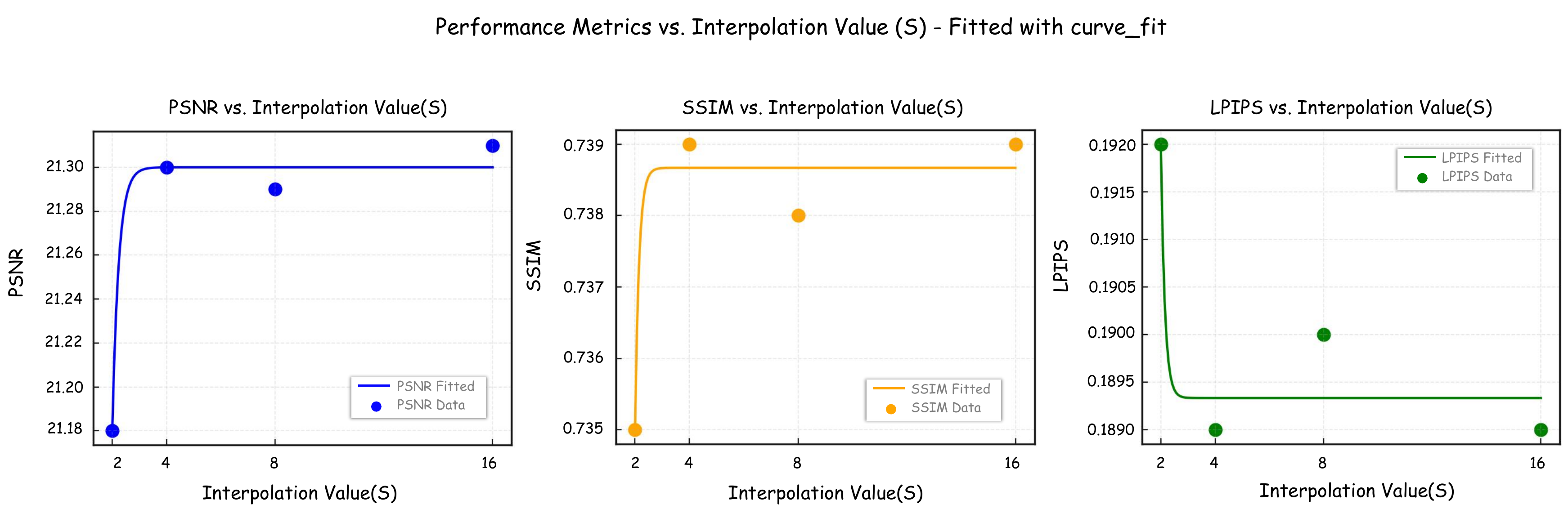}
  \caption{\textbf{ Ablation study on LLFF dataset showing the impact of varying interpolation factor $S$ on PSNR, SSIM, and LPIPS.}
  }
  \label{fig:Inter_factor}
\end{figure*}

\textbf{Ablation on Selection Module.}
Table~\ref{tab:Ablation_selection} demonstrates that incorporating the Selection module yields consistent improvements across all evaluation metrics, this improvement verifies the effectiveness of filtering out low-quality pseudo-labels, which stabilizes supervision and prevents the propagation of noise during optimization. In essence, the Selection module ensures that only reliable pseudo-labels contribute to training, leading to cleaner gradients and more robust convergence under sparse-view settings.
\begin{table}[t]
\centering

  \begin{tabularx}{0.5\textwidth}{l|*{3}{Y}}

  \hline
  Settings & PSNR & SSIM & LPIPS\\
  \hline
  w/o Selection      & 21.19 & 0.736 &  0.190\\
  All     & \textbf{21.30} & \textbf{0.739} & \textbf{0.189}\\
  \hline
  
  \end{tabularx}
  \caption{Ablation study on Selection module on LLFF dataset for 3 training views. }
  \label{tab:Ablation_selection}
\end{table}

\begin{figure*}[t]
  \centering
  \includegraphics[width=0.91\textwidth]{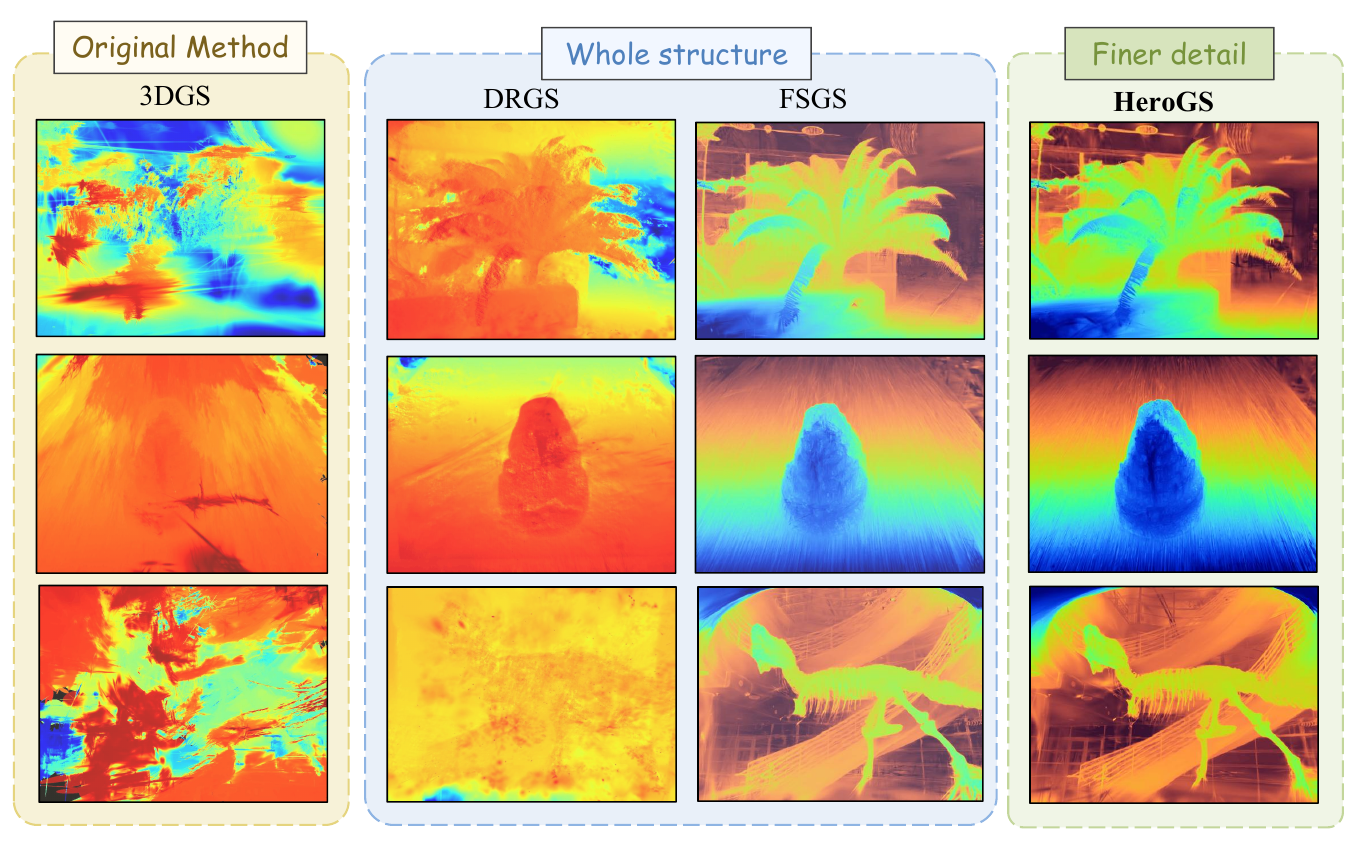}
  \caption{
  \textbf{Qualitative comparison of depth rendering quality.}
  We visualize the depth maps rendered from the reconstructed 3D Gaussian fields of different methods and our proposed HeroGS framework. 
  The visualizations show that our method produces more consistent and artifact-free depth results with finer geometry structure preservation. 
  }

  \label{fig:depth_vis}
\end{figure*}

\textbf{Depth Visualization.}
Fig.~\ref{fig:depth_vis} presents qualitative comparisons of depth maps rendered from Gaussian fields reconstructed by 3DGS, DRGS, FSGS, and our proposed HeroGS framework. 3DGS suffers from severe artifacts and structural inconsistencies, particularly near object boundaries and occlusions. DRGS improves upon this with depth supervision, yet still exhibits oversmoothing and background bleeding, compromising geometric fidelity.

FSGS enhances robustness via Proximity-guided Gaussian Unpooling, capturing global shapes more reliably but still fails to preserve finer geometric details, leading to blurry depth transitions. In contrast, HeroGS framework yields significantly sharper and cleaner depth maps. It better preserves thin structures—such as plant stems and insect limbs—and maintains clear depth discontinuities. The consistent performance gains can be ascribed to the multi-level hierarchical guidance and the synergistic coupling across different supervision levels, which jointly refine Gaussian distributions for more robust reconstruction.

\textbf{Mip-NeRF360.}
To further demonstrate the generalizability of HeroGS in unbounded real-world scenes, it is evaluated on the Mip-NeRF360 dataset using 12 and 24 input views with resolutions downsampled by 8$\times$. As shown in Tab.~\ref{tab:QUA_Mip}, HeroGS consistently achieves the best performance across all metrics. In particular, it surpasses the second-best method by a notable margin of +0.5 dB in PSNR and +0.02 in SSIM on average, while maintaining competitive perceptual quality in terms of LPIPS. These results highlight that HeroGS can effectively handle complex illumination and large-scale geometry, showing superior robustness under sparse-input conditions and strong generalization to unbounded scene reconstruction.

\section{Discussion}
\textbf{Another View of the Overall Framework.}
In our framework, a dense set of RGB images is synthesized as pseudo-labels, which, together with the training views, jointly constrain the optimization of the Gaussian Splatting field. To mitigate the potential 3D geometric inconsistencies introduced by the primary pseudo-labels, we design a refinement pipeline that incorporates two synergistic submodules. First, the Feature-Adaptive Densification and Pruning (FADP) enhances features discriminability using training views through adaptive densification controlling, while stochastically pruning redundant textures to prevent overfitting to label noise. This process encourages finer representation of geometry and textures in high-frequency regions. Second, the Co-Pruned Geometry Consistency (CPG) adopts a freeze-and-co-prune strategy to suppress erroneous structures arising from pseudo-label supervision, thereby mitigating distortion artifacts and improving global consistency. Pseudo-labels generation and two refinement submodules form a coherent framework that not only enhances supervision quality but also preserves structural fidelity under sparse-view conditions.

\textbf{Limitation and Future Work.}
Although our framework forms a hierarchical guidance across multiple levels, the current implementation only applies the Feature-Adaptive Densification and Pruning (FADP) once during the training phase. 
This design is motivated by the observation that a single round of FADP is sufficient to recover most high-frequency geometric details, whereas additional iterations only bring marginal gains in performance.
In future work, we plan to extend this mechanism into a multi-stage adaptive refinement process, where density control can be repeatedly guided by the pseudo-labels from the image level. 
Such a recurrent optimization loop would enable dynamic feedback across levels, thereby further enhancing geometric precision and strengthening global consistency in sparse-view reconstruction tasks.

\end{document}